\documentclass[acmsmall,screen,manuscript]{acmart}
\AtBeginDocument{%
  }

\setcopyright{acmlicensed}
\copyrightyear{2018}
\acmYear{2018}
\acmDOI{XXXXXXX.XXXXXXX}
\usepackage{graphicx}
\usepackage{xspace}
\usepackage{todonotes}
\newcommand{\bph}{\ensuremath{\textsc{BharatPotHole}}\xspace}
\newcommand{\rw}{\ensuremath{\textsc{iWatchRoad}}\xspace}
\newcommand{\rwn}{\ensuremath{\textsc{iWatchRoadv2}}\xspace}
\usepackage{booktabs}
\usepackage{array}
\usepackage{xcolor}
\usepackage{adjustbox}
\usepackage{pifont}
\usepackage{tcolorbox}
\newcommand{\cmark}{\textcolor{black}{\ding{51}}}  
\newcommand{\xmark}{\textcolor{red}{\ding{55}}}    

\definecolor{darkgreen}{rgb}{0.0, 0.5, 0.0} 

\acmJournal{JACM}
\acmVolume{37}
\acmNumber{4}
\acmArticle{111}
\acmMonth{8}

\begin{document}

\title{iWatchRoadv2: Pothole Detection, Geospatial Mapping, and Intelligent Road Governance}

\author{Rishi Raj Sahoo}
\email{rishiraj.sahoo@niser.ac.in}
\orcid{0009-0001-4135-6359}
\affiliation{%
  \institution{National Institute of Science Education and Research (NISER), An OCC of HBNI}
  \city{Bhubaneswar}
  \state{Odisha}
  \country{India}
}

\author{Surbhi Saswati Mohanty}
\authornote{This work was carried out while the co-author was an intern at NISER.}
\orcid{0009-0005-8533-9795}
\email{cen.22bced69@silicon.ac.in}
\affiliation{%
  \institution{Silicon Institute of Technology}
  \city{Bhubaneswar}
  \state{Odisha}
  \country{India}
}

\author{Subhankar Mishra}
\email{smishra@niser.ac.in}
\orcid{0000-0002-9910-7291}
\affiliation{%
  \institution{National Institute of Science Education and Research (NISER), An OCC of HBNI}
  \city{Bhubaneswar}
  \state{Odisha}
  \country{India}
}

\renewcommand{\shortauthors}{Sahoo et al.}


\begin{abstract}
 Road potholes pose significant safety hazards and maintenance challenges, particularly on India's diverse and under-maintained road networks. This paper presents \rwn \footnote{This is an extended version of "iWatchRoad: Scalable Detection and Geospatial Visualization of Potholes for Smart Cities"}, a fully automated end-to-end platform for real-time pothole detection, GPS-based geotagging, and dynamic road health visualization using OpenStreetMap (OSM). We curated a self-annotated dataset of over 7,000 dashcam frames capturing diverse Indian road conditions, weather patterns, and lighting scenarios, which we used to fine-tune the Ultralytics YOLO model for accurate pothole detection. The system synchronizes OCR-extracted video timestamps with external GPS logs to precisely geolocate each detected pothole, enriching detections with comprehensive metadata including road segment attribution and contractor information managed through an optimized backend database. \rwn \footnote{under-review} introduces intelligent governance features that enable authorities to link road segments with contract metadata through a secure login interface. The system automatically sends alerts to contractors and officials when road health deteriorates, supporting automated accountability and warranty enforcement. The intuitive web interface delivers actionable analytics to stakeholders and the public, facilitating evidence-driven repair planning, budget allocation, and quality assessment. Our cost-effective and scalable solution streamlines frame processing and storage while supporting seamless public engagement for urban and rural deployments. By automating the complete pothole monitoring lifecycle, from detection to repair verification, \rwn enables data-driven smart city management, transparent governance, and sustainable improvements in road infrastructure maintenance. The platform and live demonstration are accessible at \url{https://smlab.niser.ac.in/project/iwatchroad}.
 \end{abstract}



\begin{teaserfigure}
  \includegraphics[width=\textwidth]{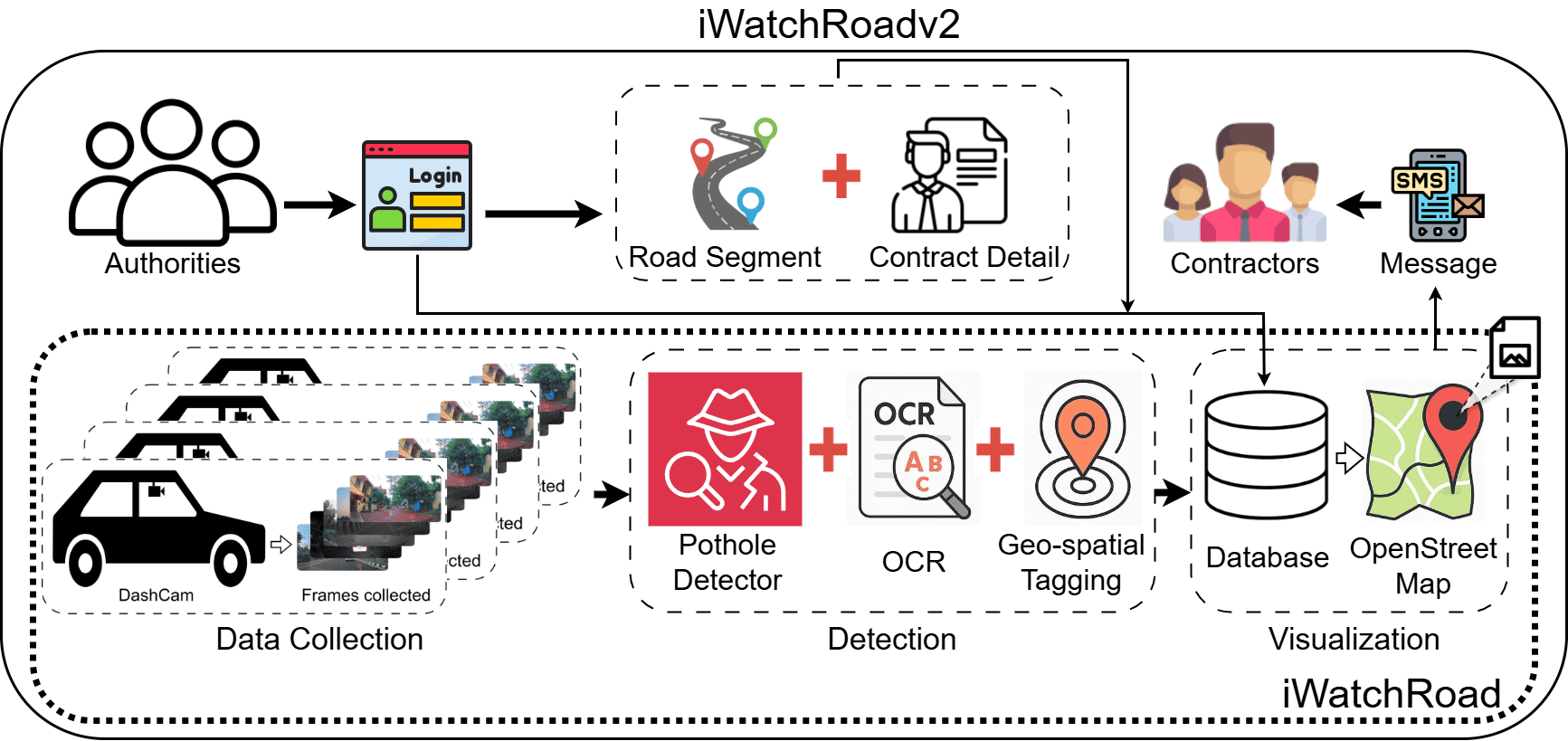}
  \caption{End-to-end \rwn pipeline: from dashcam data collection and automated pothole detection (with OCR and GPS tagging) to real-time visualization and governance using integrating contracted road metadata and seamless notification via a public OpenStreetMap interface.}
  \Description{}
  \label{fig:teaser} 
\end{teaserfigure}

\received{20 February 2007}
\received[revised]{12 March 2009}
\received[accepted]{5 June 2009}

\maketitle
\section{Introduction}
Road infrastructure maintenance plays a vital role in ensuring transportation safety and encouraging economic growth. Among various pavement deteriorations, potholes represent a particularly hazardous and prevalent road defect. These road irregularities increase the risk of accidents and contribute to spinal trauma from abrupt jolts of the vehicle. Vehicle components experience accelerated degradation and reduced operating life due to repeated hard braking required to avoid a pothole. The severity of this problem in India is evidenced by the statistics of the Ministry of Road Transport and Highways (MoRTH) documenting 2,140 and 1,471 fatalities attributed to pothole-related accidents in 2019 and 2020, respectively \cite{11031494}. Rapid identification and remediation of potholes represents a critical requirement to protect road users and preserve the integrity of the infrastructure \cite{Safyari2024}. Despite these pressing needs, municipal authorities in cities such as Pimpri Chinchwad (PCMC, Maharashtra) and several Gujarat municipalities continue to rely on manual assessment methods and citizen-initiated complaints, approaches that prove both time consuming and resource intensive. This situation underscores the urgent need for automated real time monitoring systems capable of streamlining road condition assessment and enabling rapid repair interventions.

Contemporary developments in computer vision and deep learning methodologies have demonstrated considerable potential for automated pothole identification. Numerous investigations have developed custom-trained models for detecting potholes in street-level imagery. However, significant limitations characterize existing approaches: many depend on constrained datasets comprising approximately 1,500 images, while others employ datasets from foreign contexts that inadequately capture the diverse characteristics of Indian roads, particularly rural routes, inadequately illuminated areas, and weather-impacted surfaces. Although certain systems incorporate GPS-based localization and cartographic visualization, their scope typically remains limited to urban thoroughfares or forest access roads, lacking comprehensive contextual metadata and detailed visual documentation. Additionally, previous research frequently overlooks the inclusion of negative training samples (such as utility covers and shadow artifacts) essential for minimizing false detections, and often inadequately addresses temporal synchronization challenges between video frames and positioning data.

Previous methods often failed to achieve effective geospatial integration or encountered difficulties when dealing with the diverse nature of Indian roads. In contrast, the solution proposed here offers an end-to-end approach that starts with capturing dashcam video and culminates in real-time visualization via a web platform, as illustrated in Figure \ref{fig:teaser}. The adoption of dashcams has increased nationwide, with approximately 7.2 million units sold in India during FY 2024. This ubiquity allows the system to function autonomously as participants can provide data passively without any additional input. To overcome the lack of geotagging seen in previous research, the approach aligns timestamps extracted from video frames using Optical Character Recognition (OCR) \cite{easyocr} with GPS data, guaranteeing precise location information for each detected pothole. Furthermore, the YOLO model developed \cite{redmon2016you} is trained and fine tuned on a broad, self collected Indian road dataset that spans a variety of road types, weather and lighting conditions, addressing the issue of domain mismatch and low light failures common in global datasets. The training process also incorporates non-pothole road images, which helps the model ignore misleading cues such as shadows or manhole covers, reducing false positives. Furthermore, the system uploads comprehensive and metadata rich results, including timestamps, GPS coordinates, severity levels, and detailed contractor information, to a dynamic interface built upon OpenStreetMap (OSM) \cite{openstreetmap}. This platform offers transparent, scalable, and fully actionable insights accessible to both government officials and the general public, expanding the utility of the system far beyond the mere detection of potholes. Importantly, the database is continuously updated to reflect current conditions, ensuring that when potholes are repaired, changes are accurately recorded, and the map reflects the latest infrastructure status, facilitating effective maintenance tracking and accountability.

\subsection*{Addressing Road Infrastructure Accountability Gaps}

Traditional road monitoring systems offer limited visibility into the contractor's accountability and maintenance responsibility. Existing government portals such as the Pradhan Mantri Gram Sadak Yojana (PMGSY) Online Monitoring System (OMMS) provide project level data including budgets, contractors, and sanction orders, but lack precise geographical segmentation tied to construction metadata. This creates a critical gap where road defects cannot be systematically traced to the responsible parties, hindering effective accountability and timely repairs. Our approach addresses this limitation by introducing geospatial segment mapping that directly links citizen-reported defects to specific contractors and warranty periods, enabling true data-driven accountability in infrastructure maintenance.

\subsection*{Enhanced Governance Integration}
Beyond traditional detection capabilities, our system, \rwn an advanced version of \rw \cite{sahoo2025iwatchroadscalabledetectiongeospatial}, introduces sophisticated contractor performance monitoring through the Smart Road Construction Mapper. This innovative component enables authorized personnel to securely input contractor details, construction dates, and budget information through dedicated authority logins. The system employs automated notification mechanisms using SMS triggers and email alerts to ensure accountability: when road segments transition to damaged states (indicated by color-coded visualization), simultaneous notifications are dispatched to both authorities and responsible contractors. This creates a closed-loop accountability system in which repairs can be verified through subsequent dashcam data uploads, automatically updating segment status, and maintaining comprehensive audit trails for public transparency.
 
\subsection*{Contribution}
\begin{enumerate}
    \item \bph: We introduce an extensive, self-labeled dataset that features various infrastructure of Indian roads, encompassing varied road classifications, meteorological conditions, and illumination scenarios representative of authentic driving contexts.
    
    \item Specialized Detection Architecture: We developed and publicly released a deep learning-based pothole identification model utilizing YOLO, optimized on the \bph dataset, showing strong performance in challenging operational environments.
    
    \item OCR-driven GPS Alignment: We established synchronization between dashcam video sequences and external GPS trajectories through OCR-based timestamp extraction from frames, facilitating accurate geolocation of potholes.
    
    \item Web based Mapping System: We constructed a web interface using OpenStreetMap that visualizes identified potholes along with frame imagery and contextual metadata, generating a real-time pothole cartographic system.
    
    \item Interactive Construction Accountability Interface: We designed an interactive geospatial platform for contractor oversight, associating road segments with construction metadata, and incorporating automated alert mechanisms to strengthen governance transparency.
    
    \item Comprehensive Metadata Repository: The \rwn platform provides extensive metadata that includes an assessment of the condition of the pothole, temporal data, contractor information, and warranty status via the web interface.
\end{enumerate}

\section{Related Work}

This section reviews existing work on pothole detection, geotagging, and web-based infrastructure monitoring, positioning \rwn within the landscape of automated road assessment and civic engagement platforms as shown in table \ref{tab:comparison}.
\begin{table*}[htb]
  \small
  \centering
  \caption{Comparative Analysis of Pothole Detection Systems}
  \label{tab:comparison}
  \begin{adjustbox}{max width=\textwidth}
  \begin{tabular}{@{}p{2.8cm}cccccccccc@{}}
    \toprule
    \textbf{Paper/System} & \textbf{Indian} & \textbf{Dataset} & \textbf{Pothole} & \textbf{Grading} & \textbf{Geo-} & \textbf{Meta-} & \textbf{Road} & \textbf{Contract} & \textbf{Message} & \textbf{Real-} \\
    & \textbf{Diverse} & \textbf{Avail.} & \textbf{Detection} & & \textbf{tagging} & \textbf{Data} & \textbf{Segment} & \textbf{Data} & \textbf{Trigger} & \textbf{Time} \\
    \midrule
    Dharneeshkar et al. \cite{9112424} & \cmark & \xmark & \cmark & \xmark & \xmark & \xmark & \xmark & \xmark & \xmark & \cmark \\
    Omar \& Kumar \cite{10.1007/s42979-024-02887-1}      & \cmark & \xmark & \cmark & \xmark & \xmark & \xmark & \xmark & \xmark & \xmark & \cmark \\
    Rasyid et al. \cite{8901626}       & \xmark & \xmark & \cmark & \xmark & \cmark & \xmark & \xmark & \xmark & \xmark & \cmark \\
    Hoseini et al. \cite{hoseini2023pothole}    & \xmark & \xmark & \cmark & \xmark & \cmark & \xmark & \xmark & \xmark & \xmark & \cmark \\
    SafeDrive \cite{11042541}          & \xmark & \xmark & \cmark & \xmark & \cmark & \xmark & \xmark & \xmark & \xmark & \cmark \\
    SeeClickFix \cite{seeclickfix2024}        & \xmark & \xmark & \xmark & \xmark & \cmark & \cmark & \xmark & \xmark & \cmark & \xmark \\
    FixMyStreet \cite{fixmystreet2024}         & \xmark & \xmark & \xmark & \xmark & \cmark & \cmark & \xmark & \xmark & \cmark & \xmark \\
    StreetBump \cite{Carrera2013StreetBump}            & \xmark & \xmark & \xmark & \xmark & \cmark & \xmark & \xmark & \xmark & \xmark & \cmark \\
    PublicSense \cite{wang2016publicsense}     & \xmark & \xmark & \xmark & \xmark & \cmark & \cmark & \xmark & \xmark & \cmark & \xmark \\
    Huang et al. \cite{10.1145/2983323.2983886}         & \xmark & \xmark & \xmark & \xmark & \cmark & \cmark & \xmark & \xmark & \cmark & \cmark \\
    Qiu et al. \cite{qiu2019crowdmapping}         & \xmark & \xmark & \xmark & \xmark & \cmark & \cmark & \xmark & \xmark & \xmark & \xmark \\
    Tempelmeier \& Demidova \cite{Tempelmeier_2022} & \xmark & \xmark & \xmark & \xmark & \cmark & \xmark & \xmark & \xmark & \xmark & \xmark \\
    Truong et al. \cite{truong_et_al:LIPIcs.GISCIENCE.2018.61}     & \xmark & \xmark & \xmark & \xmark & \cmark & \xmark & \xmark & \xmark & \xmark & \xmark \\
    Lincy et al. \cite{Lincy}        & \cmark & \xmark & \cmark & \xmark & \xmark & \xmark & \xmark & \xmark & \cmark & \cmark \\
    Yebes et al. \cite{Yebes_2021}     & \xmark & \xmark & \cmark & \xmark & \xmark & \xmark & \xmark & \xmark & \xmark & \cmark \\
    \rw \cite{sahoo2025iwatchroadscalabledetectiongeospatial} & \cmark & \cmark & \cmark & \cmark & \cmark & \cmark & \xmark & \xmark & \xmark & \cmark \\
    \midrule
  
    \textbf{\rwn}      & \cmark & \cmark & \cmark & \cmark & \cmark & \cmark & \cmark & \cmark & \cmark & \cmark \\
    \bottomrule
  \end{tabular}
  \end{adjustbox}
\end{table*}
\subsection{Vision-based Pothole Detection}
Computer vision techniques for pothole detection have advanced significantly in recent years. Although traditional methods such as thresholding and image segmentation were initially used, contemporary research primarily uses deep neural networks \cite{Ma_2022}. In particular, object detection convolutional neural networks (CNNs) such as Faster R-CNN, SSD, and YOLO have demonstrated high accuracy and speed, making them suitable for real-time pothole identification. For example, the study in \cite{9112424} compiles a dataset of 1500 images from Indian roads and trained models such as YOLOv3, YOLOv2, and YOLOv3-tiny to detect potholes effectively. Similarly, the research described in \cite{10.1007/s42979-024-02887-1} evaluated YOLOv5, YOLOv6, and YOLOv7 in pothole images of various Indian roadways and found YOLOv7 to achieve the highest precision. These findings corroborate that modern CNN-based detectors, particularly YOLO architectures, are highly capable of detecting road defects. Alternative vision-based approaches include stereoscopic and 3D imaging methods; stereo cameras and LiDAR can reconstruct the geometry of the road surface to find potholes, although these techniques tend to be expensive and technically complex \cite{8577119}. Other methods utilizing mobile sensors such as accelerometers or ultrasonic devices have been explored for pothole detection on the move, but these lack the capacity to provide detailed spatial localization.

\subsection{Mapping and Geotagging Potholes}
GPS-based location tagging has been widely used in road infrastructure monitoring. Research shows that GPS coordinates can be accurately extracted from dashcam recordings and effectively integrated with pothole detection frameworks \cite{8901626}. For example, the study in \cite{hoseini2023pothole} developed a deep learning-based system to detect and track potholes specifically on forest roads using dashcam footage. This system leverages object detection and tracking techniques to geolocate potholes and generate maps useful for maintenance scheduling. Additionally, SafeDrive \cite{11042541} is a well-structured platform that employs YOLOv7 trained on foreign datasets combined with cloud technologies for pothole detection and mapping. However, these existing systems often lack sufficient data diversity from Indian roads and detailed information on infrastructure specific potholes.

\subsection{Web-based Crowdsourcing and Civic Engagement Platforms}
The intersection of web technologies, crowdsourcing, and monitoring of urban infrastructure has received significant attention in web science research. Crowdsourcing systems on the World Wide Web have transformed web platforms and created new fields of citizen participation \cite{vahdatnejad2022surveycrowdsourcingapplicationssmart}. This paradigm shift has enabled novel approaches to urban sensing and infrastructure management.

\subsubsection{Crowdsourcing for Urban Sensing}
Web based crowdsourcing platforms \cite{doan2011crowdsourcing} have emerged as powerful tools to collect urban data on a scale. Crowdsourcing based urban anomaly reporting systems enable pervasive and real-time reporting of anomalies in cities, including noise, illegal use of public facilities, and urban infrastructure malfunctions \cite{10.1145/2983323.2983886}. Research has explored predictive systems that leverage crowdsourced data to anticipate urban anomalies before they become critical issues. These systems typically rely on explicit citizen reports rather than automated detection mechanisms. The concept of "citizens as sensors" has been extensively studied in geospatial web research. Wang et al. \cite{wang2016publicsense} developed PublicSense, a crowdsensing platform for public facility management in smart cities, demonstrating how distributed citizen participation can supplement traditional monitoring approaches. However, these platforms face challenges with duplicate reporting, data quality validation, and sustained citizen engagement.

\subsubsection{Collaborative Mapping and OpenStreetMap Research}
OpenStreetMap (OSM), as a collaborative crowdsourced web mapping platform, has been the subject of extensive research in web conferences. OSM represents a unique source of openly available worldwide map data that is increasingly adopted in web applications, though vandalism detection remains a critical task to support trust and maintain transparency. This research has focused on attention-based approaches to detect and prevent vandalism in collaborative mapping environments \cite{Tempelmeier_2022}, highlighting the challenges of maintaining data quality in crowdsourced geospatial platforms. Earlier work by Truong et al. \cite{truong_et_al:LIPIcs.GISCIENCE.2018.61} explored cluster based outlier detection methods for identifying vandalism in OSM, while comprehensive reviews have analyzed emerging trends in vandalism detection techniques.

Crowd mapping urban objects from street-level imagery \cite{qiu2019crowdmapping} has been explored in Web Conference research, improving public transit accessibility by crowdsourcing landmark locations with Google Street View. This work demonstrates the potential of combining computer vision with crowdsourced validation to enhance urban infrastructure databases. However, these approaches typically require manual annotation and lack automated real-time detection capabilities.

\subsubsection{Citizen Reporting Platforms}
Traditional citizen reporting platforms have established mechanisms for municipal issue reporting. SeeClickFix \cite{seeclickfix2024} provides a comprehensive 311 request and work management system that allows residents to report various municipal problems, including potholes, graffiti, and broken street lights. FixMyStreet \cite{fixmystreet2024}, developed by mySociety, offers an open source platform that allows citizens to report street problems by marking them on an interactive map. These platforms have been deployed globally and facilitate direct communication between citizens and local authorities.

Mobile-first approaches have simplified the reporting process. Applications like Stan The App and the 311 Universal platforms enable citizens to capture and transmit images along with GPS coordinates. Research on crowdsourcing based Android platforms \cite{6975489} has explored selective connectivity between citizens and governing bodies for sharing public views on road conditions. The StreetBump application \cite{Carrera2013StreetBump} represents a pioneering example of "subliminal" crowdsourcing that automatically collects road condition information using smartphone accelerometers and GPS without human intervention. Systems that integrate deep learning with business process management platforms like Camunda have been proposed to visualize and track pothole maintenance workflows.

\subsubsection{Web Technologies for Smart Cities}
The ACM Web Conference has hosted workshops on "The Responsible Web and AI for Smart Cities" (WebAndTheCity) \cite{anthopoulos2024webandthecity}, recognizing the growing importance of web technologies in urban governance. Research in this domain explores how web platforms can foster transparency, accountability, and citizen participation in infrastructure management. However, most existing systems focus on manual reporting mechanisms rather than continuous automated monitoring.

\subsubsection{Limitations of Existing Approaches}
Although prior work has made significant contributions, several gaps remain:
\begin{itemize}
    \item \textbf{Manual vs. Automated Detection:} Most web-based reporting platforms rely entirely on manual citizen reports, requiring active user participation to identify and submit issues. This creates coverage gaps and reporting delays.
    \item \textbf{Lack of Domain-Specific Training:} Generic pothole detection models trained on foreign datasets do not generalize to various Indian road conditions, including varied surface materials, lighting scenarios, and weather patterns.
    \item \textbf{Incomplete Metadata:} Existing platforms rarely provide comprehensive temporal tracking, contractor accountability information, severity assessments, or automated repair verification mechanisms.
    \item \textbf{GPS Synchronization Challenges:} Previous dashcam based systems do not adequately address the challenge of synchronizing video timestamps with external GPS logs, leading to geolocation inaccuracies.
    \item \textbf{Limited Public Transparency:} Many systems collect data but do not provide open, accessible web interfaces for citizens and authorities to monitor road conditions and maintenance progress in real-time.
    \item \textbf{Absence of Negative Samples:} Previous detection systems often lack training on negative samples (shadows, manhole covers, tar patches), resulting in high false positive rates.
\end{itemize}

\subsubsection{Positioning \rwn}
Our work addresses these limitations by providing an end-to-end web-based system that combines:
\begin{itemize}
    \item Automated detection using region-specific YOLO fine-tuning on BharatPotHole dataset
    \item Precise geotagging through OCR-based timestamp extraction synchronized with GPS logs
    \item Public transparency via an interactive OpenStreetMap-based web interface
    \item Comprehensive metadata including severity, contractor information, and temporal tracking
    \item Repair verification through automated status updates when roads are fixed
    \item Scalable architecture supporting continuous dashcam data ingestion
\end{itemize}

Unlike traditional crowdsourcing platforms that rely on manual citizen reporting, \rwn operates automatically and continuously as dashcam-equipped vehicles traverse roads. Unlike vision-only detection systems, we integrate geospatial web technologies to provide actionable insights for governance. By bridging computer vision, web platforms, and civic engagement, \rwn advances the state-of-the-art in automated and transparent infrastructure monitoring.

\subsection*{Other Related Work}
Numerous studies focus on vehicle-mounted pothole alert systems. These smart frameworks integrate deep learning techniques with database logging to provide real-time warnings to drivers and notify authorities promptly \cite{Lincy}. Another research avenue employs thermal imaging or stereo vision methods to detect potholes under difficult conditions \cite{Yebes_2021}. Although these approaches enhance detection robustness, they still depend fundamentally on geospatial tagging to accurately locate the potholes.

\section{\bph: Curation, Diversity and Annotation}
A comprehensive dataset for pothole detection was systematically compiled using dashcam recordings obtained from various Indian road networks. The data acquisition process spanned three months (May through July), utilizing dashcam-mounted vehicles to capture images across varying road infrastructure and meteorological conditions.
The dataset incorporates multiple road typologies, from unpaved rural roads and village roads to urban arterial roads and national highways. To ensure environmental heterogeneity, footage was recorded under various climatic conditions, including monsoon precipitation and clear, dry weather. In addition, temporal diversity was achieved by capturing data across different illumination scenarios, including daylight hours, twilight periods (dawn and dusk), and nighttime conditions with artificial street lighting.

\begin{figure*}[htb]
  \centering
  \includegraphics[width=\linewidth]{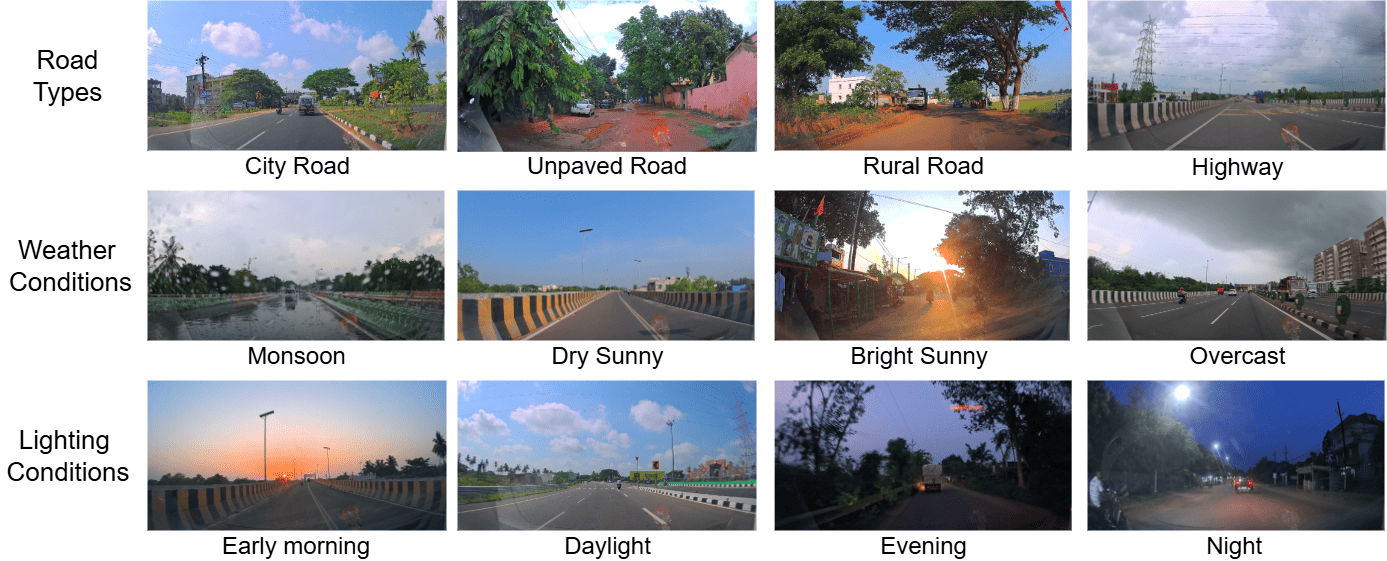}
  \caption{A comprehensive dataset covering diverse road types, weather, and lighting conditions to fine-tune the model for improved robustness and real-world performance.}
  \Description{}
  \label{fig:dtype}
\end{figure*}

Representative samples of the dataset, showing its diverse characteristics, are presented in Figure \ref{fig:dtype}. This multifaceted data collection approach is crucial for training robust pothole detection models with strong generalization capabilities across varied real-world deployment environments.

All frames exhibiting visible pothole instances underwent meticulous manual annotation, with precise bounding boxes delineated using the Roboflow platform \cite{dwyer2025roboflow}. The annotations were subsequently transformed into a YOLO format to ensure compatibility with the model training pipelines. The finalized dataset contains over 7,000 annotated frames, all sourced exclusively from our dashcam recordings, making it the most comprehensive pothole detection dataset specifically designed for Indian road infrastructure. The exclusive reliance on internally collected dashcam footage guarantees data consistency and authenticity, making the dataset particularly well suited for fine-tuning contemporary object detection architectures such as YOLOv8 \cite{yolov8_ultralytics} for deployment in Indian road environments.

The dataset adheres to conventional train-validation-test partitioning with a split ratio of 70:20:10. All images and annotations were uniformly formatted at 640×640 pixel resolution, consistent with standard practices in object detection benchmarking. The entire dataset is publicly accessible at \url{www.kaggle.com/datasets/surbhisaswatimohanty/bharatpothole} to support reproducibility and collaborative research efforts.

\subsection{Pothole Data Perspective}

\begin{figure}[ht]
  \centering
  \includegraphics[width=\linewidth]{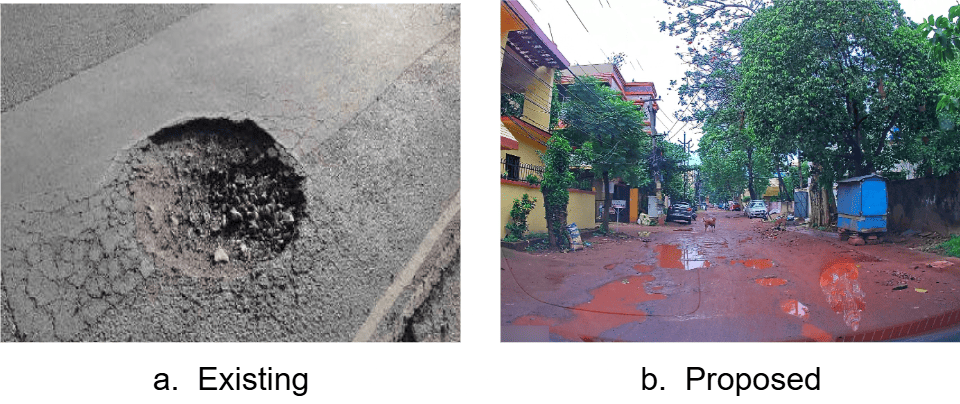}
  \caption{Comparison between traditional top-down pothole dataset (a), and the proposed forward facing dashcam view (b), which reflects real world driving scenarios and supports practical, continuous data collection for robust pothole detection.}
  \Description{}
  \label{fig:pv}
\end{figure}

The presented dataset captures roadway imagery from the operator's viewpoint, offering a feasible methodology for ongoing vehicular data collection, as illustrated in Figure \ref{fig:pv}. This forward-looking perspective aligns directly with the input received by operational pothole detection systems in authentic driving contexts. This methodology differs from conventional pothole datasets that primarily employ overhead or close-range images acquired from fixed positions situated directly above pothole locations. Although such elevated vantage points provide superior morphological information, they demonstrate limited practical applicability for vehicular applications and introduce considerable logistical complexities for sustained data gathering.

Employing dashcam recordings simplifies data acquisition by eliminating the need for specialized camera arrangements or expensive equipment deployment. However, this viewpoint presents intrinsic detection difficulties: remote potholes manifest themselves as vague dark areas on the road surface, potentially resulting in incorrect classification of harmless shadows or surface irregularities as pothole occurrences. The substantial scope and thorough annotation of the dataset facilitate robust model development to distinguish genuine pothole characteristics from misleading visual anomalies.

\subsection{Environmental Variability and Robustness}
The visibility of the hole varies considerably under environmental conditions, and readily apparent features in daylight become nearly invisible at night. Wet conditions further complicate detection through reflections and altered shadow patterns, causing models trained solely on clear day imagery to exhibit poor performance in adverse scenarios.

The environmental diversity of our dataset addresses this challenge by training the model under comprehensive operational conditions. This approach enables the development of robust feature representations that accurately detect potholes while remaining invariant to confounding factors such as shadows, reflections, and moisture. Incorporating challenging environmental conditions significantly improves system performance and reliability in a variety of deployment scenarios.
\subsection{Negative Sample Integration}

The dataset integrates comprehensive negative samples that included road images captured under various conditions and surface configurations that are devoid of pothole instances. These frames encompass various characteristics of the road surface, including fissures, cast shadows, road debris, oil stains, and manhole covers, none of which receive pothole annotations. The strategic inclusion of these negative samples is essential for mitigating false positive detections that would otherwise undermine system dependability.
\begin{figure}[ht]
\centering
\includegraphics[width=\linewidth]{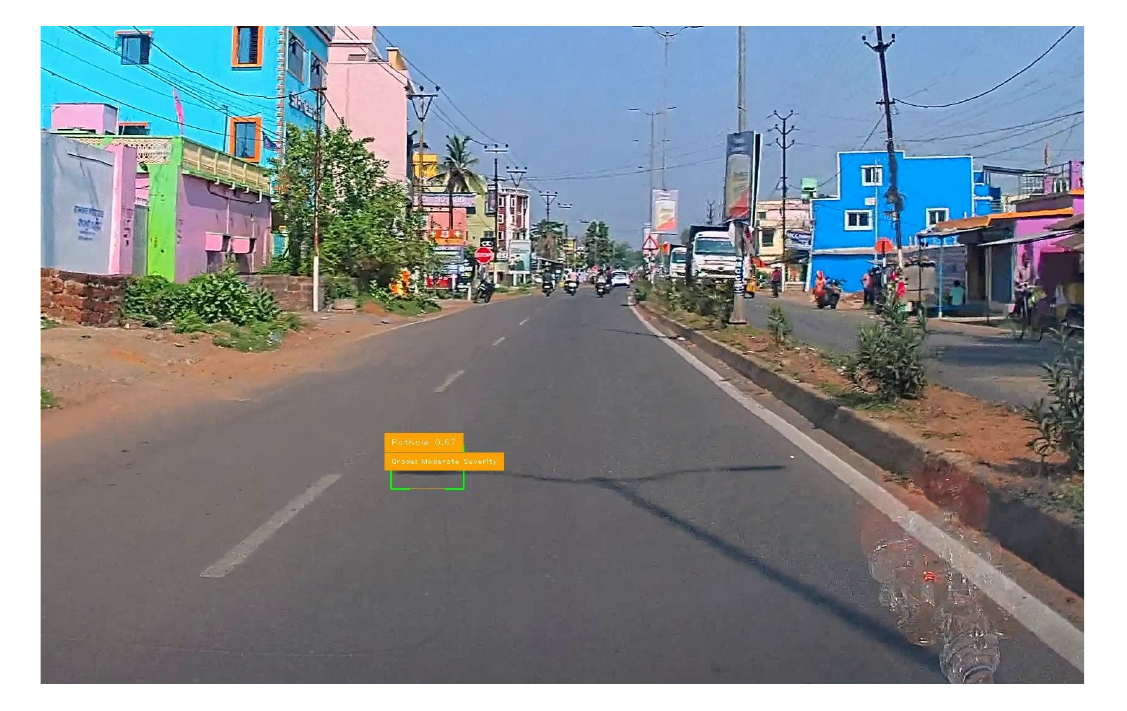}
\caption{Shadow misclassification as a pothole demonstrates the necessity of comprehensive negative sample training for robust feature discrimination.}
\Description{}
\label{fig:Lamp}
\end{figure}

In the absence of sufficient negative training examples, detection models exhibit a propensity to erroneously classify visually analogous road elements, such as utility access covers or shadows, as potholes based merely on superficial visual attributes like surface darkness. Figure \ref{fig:Lamp} exemplifies this misclassification behavior when models receive inadequate exposure of negative samples during training. The integration of these negative samples allows the learning algorithm to extract discriminative features anchored in authentic pothole characteristics, including irregular geometry, surface discontinuities, and distinctive edge profiles, rather than relying on simplistic visual indicators such as dark surface coloration.

Empirical validation revealed that models trained solely on positive pothole instances demonstrate higher false positive rates. Judicious integration of positive pothole samples with negative road surface examples promotes robust learning of pothole-specific morphological attributes (irregular depressions with degraded boundaries) as opposed to generalized recognition of dark surface pattern recognition. This balanced training strategy marks out the specificity of the model and overall detection accuracy in operational deployment scenarios.

\begin{figure*}[!htbp]
  \centering
  \includegraphics[width=\linewidth, height=6cm]{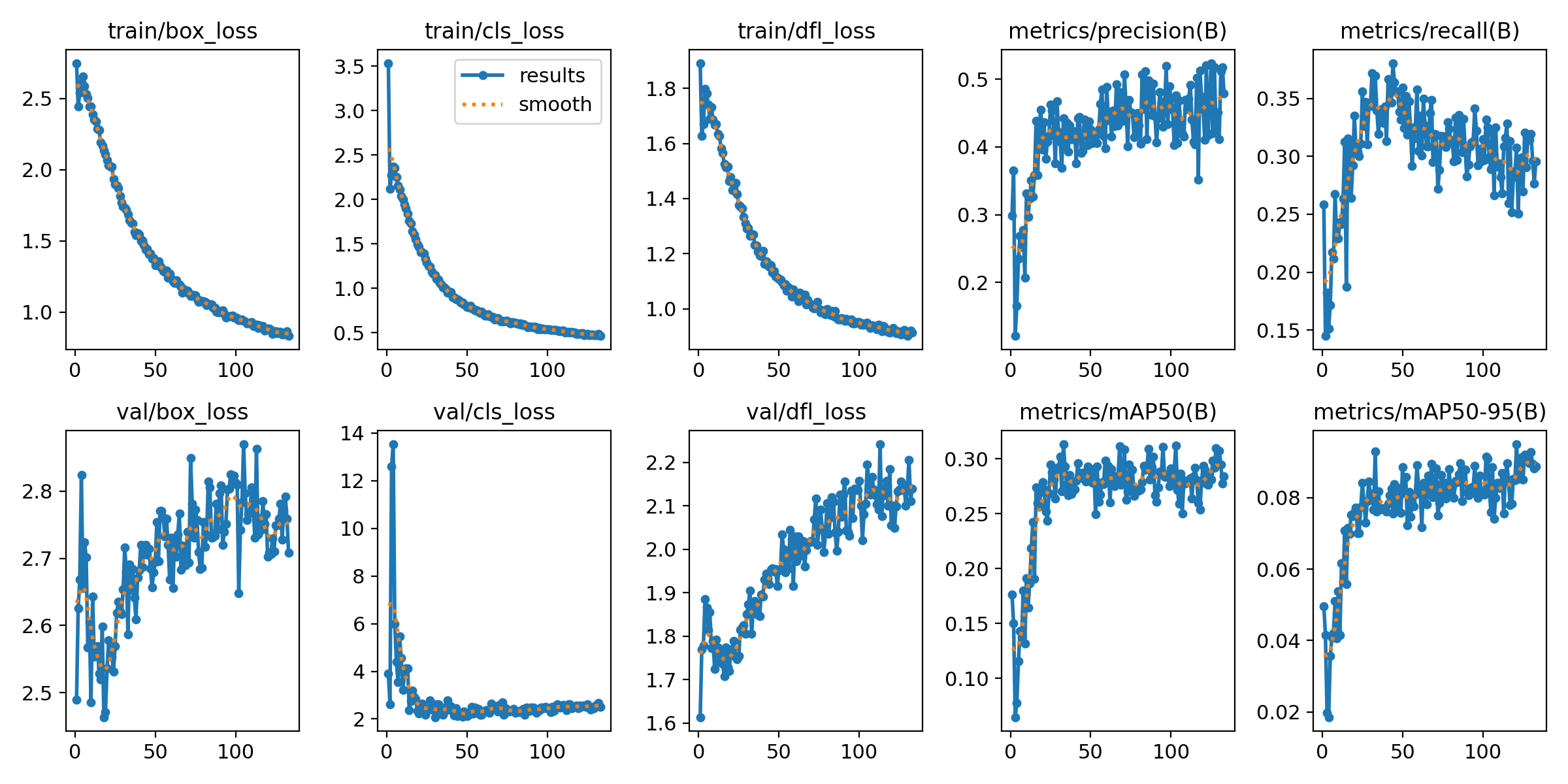}
  \caption{Model's performance on \bph-3K}
  \Description{}
  \label{fig:metric3k}
\end{figure*}
\begin{figure*}[!htbp]
  \centering
  \includegraphics[width=\linewidth, height=6cm]{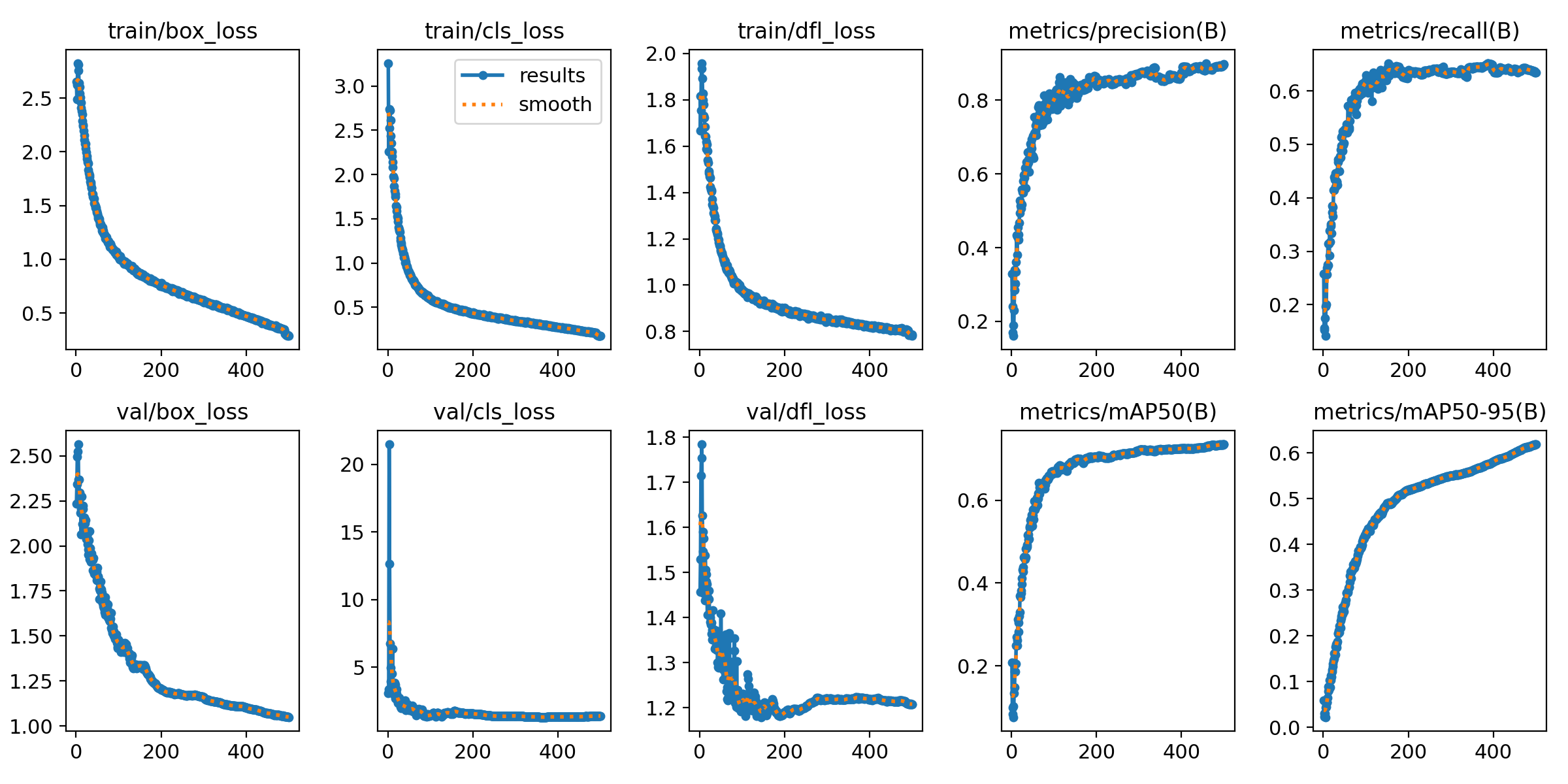}
  \caption{Model's performance on \bph-7K}
  \Description{}
  \label{fig:metric7k}
\end{figure*}

\begin{figure*}[h]
  \centering
  \includegraphics[width=\linewidth]{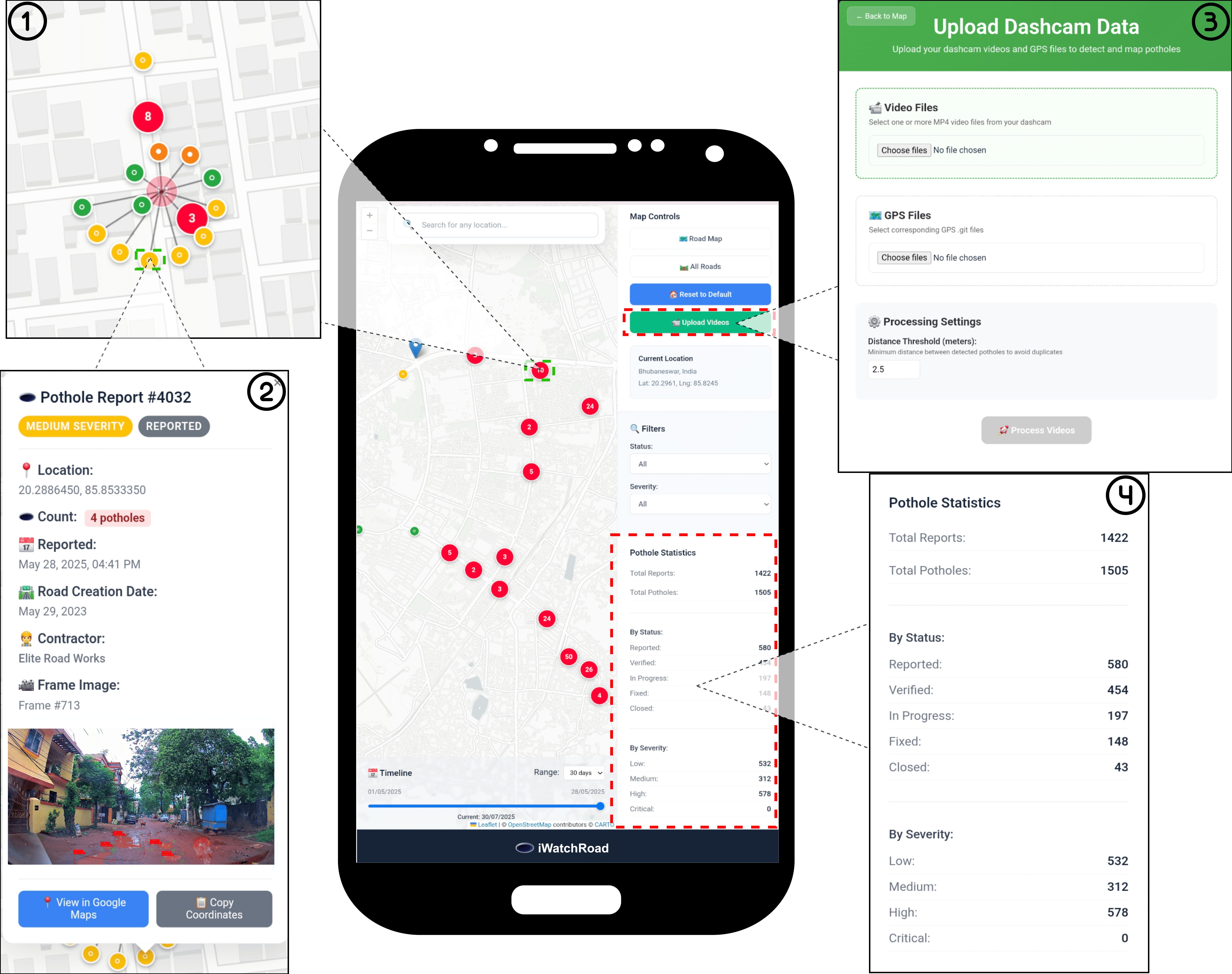}
  \caption{\rw's Web Platform Interface: Interactive map view showing geotagged pothole reports with color coding based on severity,(1) shows a cluster of potholes, and when clicked, it shows (2) detailed metadata for each pothole, including timestamp, location, and frame image. (3) Upload interface where users can submit dashcam video and GPS files for automatic pothole detection and mapping. (4) Enumerates the pothole status.}
  \Description{}
  \label{fig:web}
\end{figure*}

\section{\rw}
This section describes the architectural framework of the \rw system. Indian roads pose distinctive challenges for automated detection frameworks due to heterogeneous surface compositions, variable construction standards, and intricate meteorological patterns. The architecture manages these challenges through a modular and expandable structure.
 
\subsection{Data Collection}
Dashcam based data acquisition enables continuous dataset expansion without requiring manual intervention, as user-contributed footage is automatically aggregated to our server infrastructure. This approach significantly reduces data collection costs and operational overhead. Prior to processing, all personally identifiable information, including vehicle registration numbers and geographical addresses, is subject to automated anonymization through blurring techniques to ensure privacy protection.

\subsection{Detection}
The detection framework consists of three principal components:

\subsubsection{Custom trained YOLOv8 Detector:}
The model experiences fine-tuning employs our self-annoted dataset \bph, which contains thousands of labeled frames documenting genuine Indian road infrastructure scenarios, encompassing low-light conditions, rain-affected surfaces, and unpaved roads. The effectiveness of the model showed significant enhancement with the growth of the dataset from \bph -3k to \bph- 7k, as shown in Figures \ref{fig:metric3k} and \ref{fig:metric7k}, respectively. The fine-tuning process allows the model to distinguish authentic potholes from visually comparable road elements such as shadows, manhole covers, and road markings. The detection results produce bounding boxes surrounding identified potholes along with confidence metrics, which are then used to assess severity according to dimensional and geometric properties.

\subsubsection{Optical Character Recognition (OCR):}
EasyOCR combined with regular expression (regex) processing extracts and normalizes embedded timestamps from video frames. Because dashcam overlays frequently differ in format and placement, a specialized regex processor is utilized to restructure and sanitize the extracted text into a uniform DD-MM-YYYY HH:MM:SS format. This guarantees reliable synchronization of each frame with the associated GPS information. OCR is deployed modularly, enabling activation only on designated frames to enhance performance and resource efficiency.

\subsubsection{Geo-spatial Tagging:}
GPS recording devices persistently capture vehicle location and orientation data. Temporal alignment between GPS log records and video frame timestamps facilitates accurate geotagging of identified pothole occurrences.

\begin{figure*}[h]
  \centering
  \includegraphics[width=\linewidth]{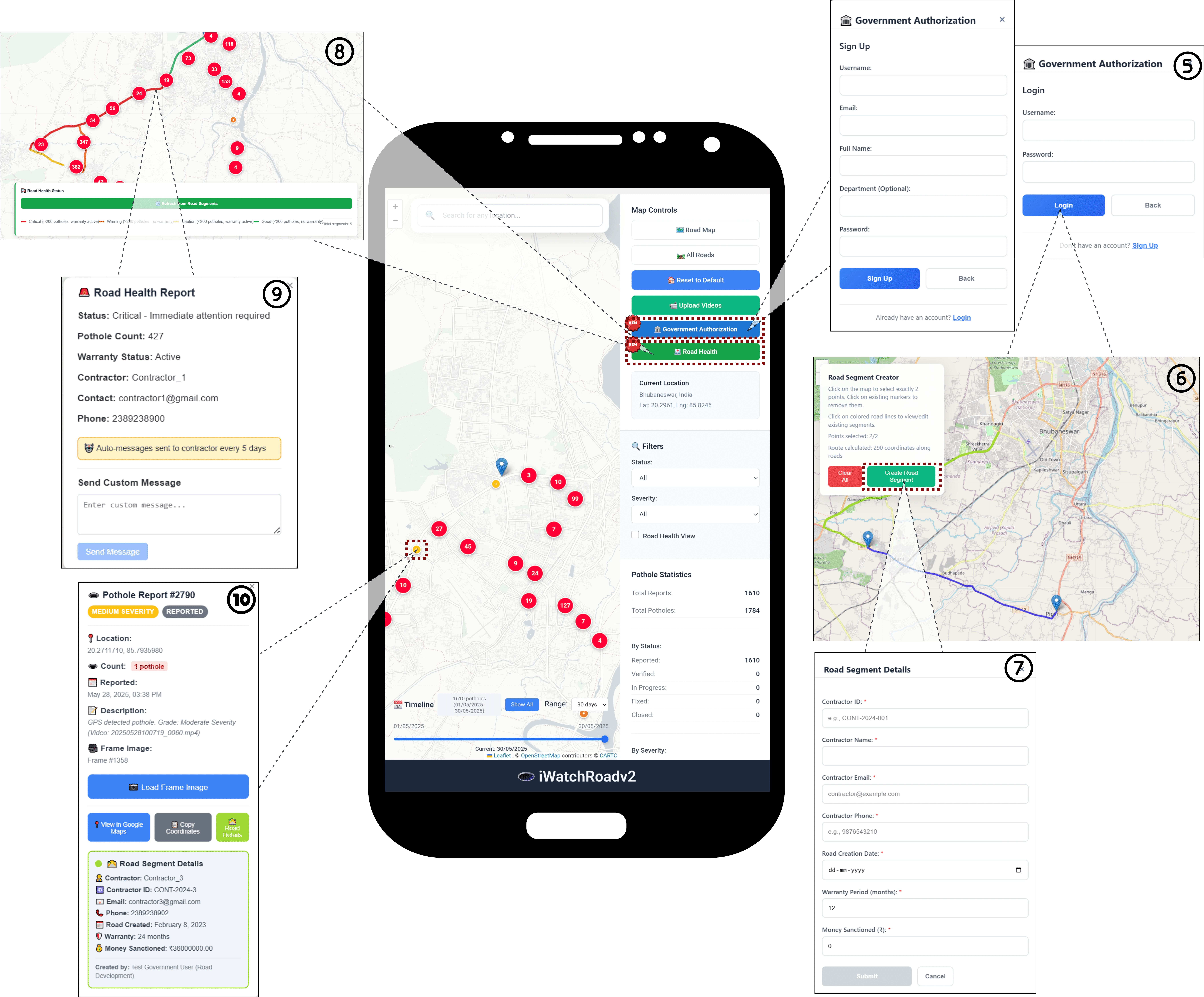}
  \caption{\rwn's Enhanced Web Platform Interface: Advanced interactive map displaying geotagged pothole reports with severity-based color coding and road segment health visualization. (5) Government Authorization module for secure authority login and contractor management access. (6) Road Segment Creator interface allowing authorities to map road segments with contractor details and project metadata. (7) Road Segment Details form capturing contractor information, construction dates, warranty periods, and budget allocations for infrastructure accountability. (8) Color-coded road segments showing health status with real-time pothole density mapping along contracted road sections. (9) Road Health Report dashboard providing critical alerts, warranty status, contractor contact information, and automated messaging capabilities for proactive maintenance management. (10) Enhanced pothole report interface with integrated road segment details, contractor information, and direct communication features for streamlined governance and accountability.}
  \Description{}
  \label{fig:webv2}
\end{figure*}

\subsection{Visualization}
This component encompasses a structured database and a web visualization tier:

\subsubsection{Database}
An SQLite database preserves metadata entries that encompass temporal data, geographic coordinates, severity categorizations, and linked frame identifiers.

\subsubsection{OpenStreetMap (OSM)}
The geotagged pothole entries are rendered on a web application constructed with OpenStreetMap utilizing Leaflet.js. Figure \ref{fig:web} displays the website's interface. Each pothole is represented by an icon on the map, indicating the number of potholes at that position. The icon, when selected, reveals a pop-up that contains metadata: a thumbnail frame of the location, detection time, pothole severity, road construction date, and contractor identity. We deploy this on our server, enabling users to navigate to any region of India and view documented potholes. Our system also supports filtering by date or road category. This platform consequently delivers a real-time pothole map: for instance, a grouping of markers signifies a severely deteriorated road requiring immediate repair. After the road is repaired and the fresh dashcam data is uploaded, the server automatically eliminates the prior marker and designates it as resolved. The website is accessible at \url{https://smlab.niser.ac.in/project/iwatchroad}.

The pipeline is straightforward; new data can be integrated with the code simply by uploading the dashcam video and the external GPS source. It will identify, geotag, upload to the database, and begin displaying on the website. This is a modular pipeline, that can be enhanced with new technological interventions in the future. The code is accessible at \url{https://github.com/smlab-niser/iwatchroad}

\section{Interactive Road Contract Mapping and Accountability}
In India, government portals such as the Pradhan Mantri Gram Sadak Yojana (PMGSY) Online Monitoring System (OMMS) and state level works department websites provide partial transparency with respect to road construction projects. These platforms typically disclose project identifiers, sanctioned road length, package numbers, executing agencies, contractor names, estimated budgets, and year of sanction. They also feature a public grievance mechanism with a complaints helpline number, allowing citizens to formally raise issues, although the actual complaint is not publicly shared. Despite these provisions, several critical attributes remain absent in structured or accessible form. Specifically, portals do not consistently report construction start and completion dates, warranty or defect liability periods for each project, or detailed information on the exact segments or chain of the roads constructed. Although such data may exist within tender documents or contractual appendices, they are rarely available in machine readable formats, thereby limiting usability for large scale analysis or integration with geospatial systems.

In the international arena, there are comparable public infrastructure platforms with varying degrees of completeness. For example, the United Kingdom’s Contracts Finder and National Highways project repositories, the United States’ USASpending.gov in conjunction with state Department of Transportation portals, and Canada’s CanadaBuys system all provide structured information on road projects, including contractor details, award dates, contract durations, and sanctioned budgets. Similarly, the European Union’s Tenders Electronic Daily (TED) and Australia’s National Infrastructure Pipeline provide structured public access to information on awarded contractors and the corresponding financial allocations for transportation projects. However, as in the Indian context, detailed attributes such as defect liability or warranty clauses and precise geographical segments of roads are often embedded within tender documents or technical appendices rather than being published as standardized open datasets. This reflects a broader challenge in public procurement transparency, where high-level contract metadata is widely available, but fine-grained operational details necessary for accountability and geospatial integration remain inaccessible.

To bridge these gaps, we introduce the Smart Road Construction Mapper, an interactive geospatial interface that allows users to mark on a map the precise segment (from-point to to-point) constructed by a particular contractor, linking each annotated section with associated project metadata. By explicitly capturing the geographic footprint of contracts, this tool not only supplements existing procurement platforms, but also enhances accountability, enables contractor-specific oversight, and facilitates integration with monitoring systems. The mapper allows intuitive interaction through a user-friendly interface, where contractors’ completed sections can be visually distinguished and queried, allowing stakeholders to trace responsibility for maintenance, delays, or recurring issues. This geospatial mapping capability has the potential to transform road project monitoring by fostering transparency, strengthening citizen engagement, and supporting data-driven policy analysis, ultimately setting a precedent for integrating digital governance with infrastructure accountability.

\subsection{Intuition}

The existing road monitoring systems in India offer limited visibility into the actual road conditions.  For example, the PMGSY portal (OMMAS/eMARG) provides high level project data – budgets, sanction orders, contractors, etc. – to promote transparency and accountability. However, these platforms report on entire road works or administrative road IDs, not on finer sections of pavement. In practice, one can see that “Road X” was built by contractor Y with so many crores sanctioned, but the interface does not locate where along the 10km stretch damage has occurred or whether the defect liability period is still in force. Moreover, this information is not easily exported or linked to GPS coordinates for automated analysis. This gap motivates our segmentation approach: by dividing each road into geospatial segments tied to official project metadata, we can directly map reported potholes (or repairs) to the responsible segment. Segment level mapping thus provides the missing bridge between citizen sourced defect reports and government records, enabling true data-driven accountability in road maintenance.

\subsection{Workflow}
Our implementation is based on OpenStreetMap (OSM) for the map interface and the Open Source Routing Machine (OSRM) routing engine for segment extraction. In the user interface, a resident or official marks a start point and an end point on the OSM map. We then call OSRM to compute the most likely road route between these two coordinates, connecting the segment to existing roads. The returned polyline is stored as a “road segment” in our database and is drawn on the map.  Each segment is fully editable: users can drag its endpoints to correct alignment, delete an incorrect segment, or update its geometry if needed. The segment record also stores all pertinent attributes for accountability; for example, as in municipal GIS systems, we include fields for the contractor’s name, the construction date, and the approved budget or sanctioned funds.

Once on the map, each segment is visualized with a dynamic color code based on the damage reported and the warranty status.  We define green segments as healthy or good (less potholes reported, although not under warranty), yellow segments as caution but repair still due (within the Defect Liability Period), orange segments as damaged (more potholes with no warranty), and red segments as severely deteriorated or past warranty. As new dashcam data are uploaded, the system counts potholes per segment and automatically updates its color.  Clicking on a segment brings up a pop-up of its metadata – for instance, similar to how our pothole markers display a mini image and info, the segment pop-up shows the construction date and contractor name. In this way, each segment of the map becomes a live repository of crowdsourced condition data and official project information.

The road monitoring system has been extended to make it more informative and interactive, thereby improving user accessibility and providing deeper insights into both road conditions and contractor performance. A dedicated authority login has been introduced, allowing authorized personnel to securely enter contractor details, road creation dates, and sanctioned budget amounts after credential verification. This ensures that the database remains accurate and up-to-date with official records.

All of these details are also made publicly accessible, as shown in Figure \ref{fig:webv2}, allowing citizens to view and follow the quality of the road and the performance of the contractors in real time. The pothole database is dynamically updated as new data is uploaded to the server through dashcam footage, ensuring that the system reflects the most current ground conditions. By integrating geospatial visualization, automated alerts, and open accessibility, the platform strengthens transparency, fosters accountability, and supports proactive maintenance of road infrastructure.

\subsection{Repair Verification and Trigger Mechanism}
As soon as a segment status turns yellow or red, \rw automatically notifies the responsible parties.  For example, a segment that first turns yellow (indicating a report of caution under warranty) will trigger an alert email or app notification to both the contractor and the local road authority. If the segment is not fixed by the warranty deadline, the status flips to red and the system escalates the message to higher officials (for example, senior engineers or the public works department). This rule based alert mechanism ensures that delayed repairs are flagged and that contractual obligations are enforced.

We also implement an automatic verification loop to confirm repairs. Once a segment has been marked for repair, the next time a user traverses that road segment, the system checks the new incoming data. In practice, if the dashcam on a subsequent trip shows that the previously reported potholes no longer appear, the segment is considered fixed.  At that point, the segment status is cleared: the warning marker is removed (or turned green), and the database entry is updated to 'repaired'. This mimics the logic in our pothole pipeline: "once the road is repaired and new dashcam data is uploaded, the server automatically removes the previous marker and marks it as fixed". In this way, \rwn closes the feedback loop; citizen data not only identify defects but also verify that maintenance has been completed.
\begin{figure*}[htbp]
  \centering
  \includegraphics[width=\linewidth]{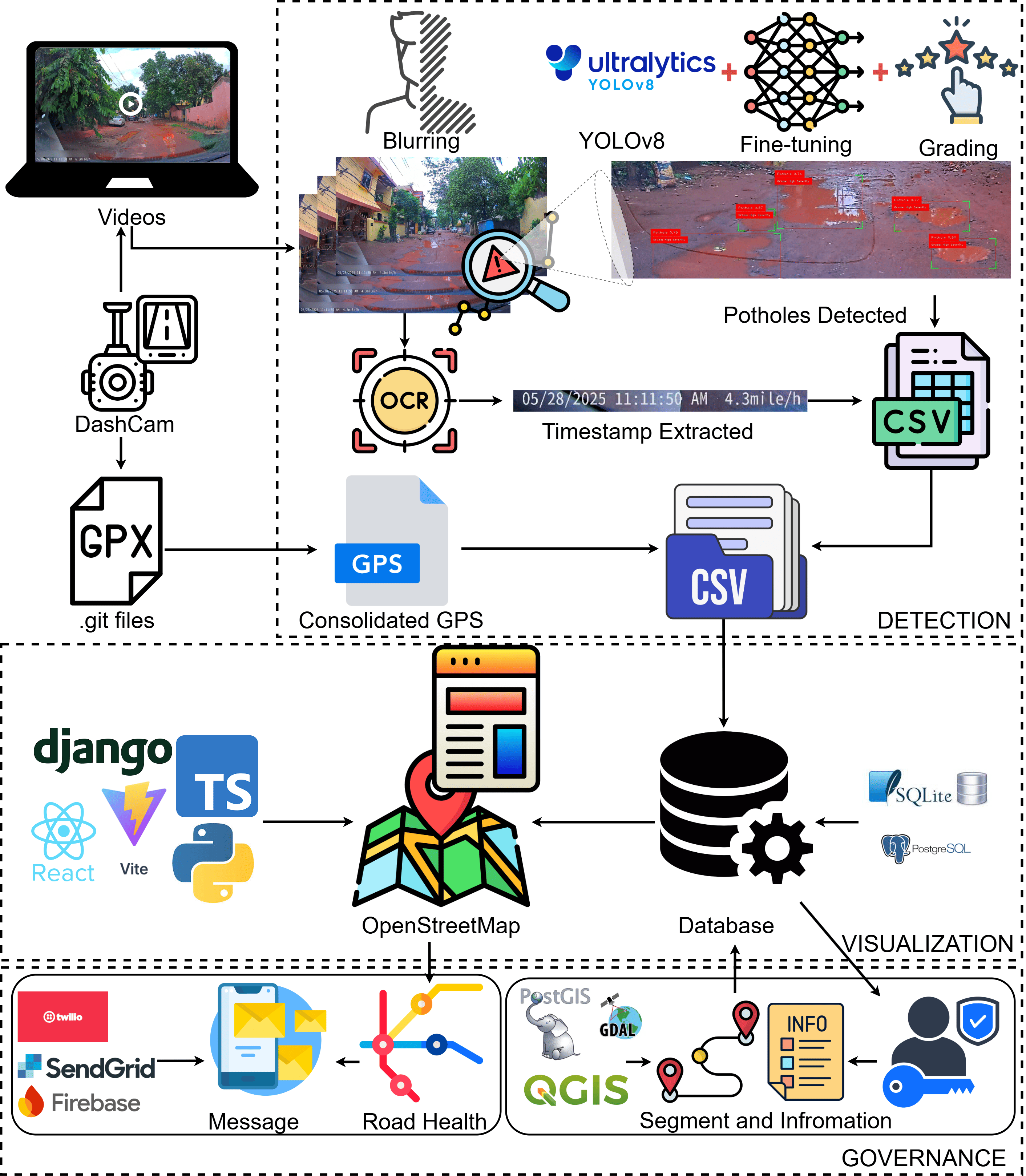}
  \caption{The diagram illustrates the complete workflow of our automated pothole detection and road governance platform. Dashcam videos are segmented into frames, and privacy is safeguarded by blurring sensitive details. A fine-tuned YOLOv8 model detects potholes, while OCR extracts timestamps synchronized with GPS data for accurate geolocation. Detection results and metadata are stored in CSV files and managed in a database, powering an interactive web interface built with Django, React, and TypeScript, and visualized via OpenStreetMap. The system also features integrated road health analytics, segment-specific contract data, automated notification (Twilio, SendGrid, Firebase), and secure authority authentication, enabling transparent infrastructure oversight and proactive road maintenance.}
  \Description{}
  \label{fig:roadwatch}
\end{figure*}

\section{Experiment}
The online pothole detection and road governance platform \rwn was extensively evaluated using a complete automated pipeline encompassing real-world road data collection, AI-based detection, advanced geospatial analytics, enriched database integration, authority authentication and an interactive public facing web application. The experiment assessed the system’s effectiveness in seamlessly detecting potholes, delivering precise geotagging, supporting contractor accountability, and presenting actionable insights to both road authorities and citizens through a scalable interface.

\subsection{Experiment Setup}

System validation was performed on standard consumer hardware, highlighting both affordability and accessibility. The end-to-end solution performed efficiently on typical multi-core Intel i5/i7 CPUs with 16 GB RAM; high-end RTX 3090 GPUs were used solely for model training and fine-tuning. The backend was implemented in Django (Python 3.11+), featuring robust user authentication and authority management modules, while the frontend leveraged React and TypeScript with Vite for rapid, responsive web performance. Leaflet.js and React-Leaflet powered dynamic geospatial visualization; OpenStreetMap formed the underlying mapping foundation. Road health analytics and segment-level contract data were managed using PostGIS and QGIS integrations, further supporting spatial queries and infrastructure metadata.  This technology stack guarantees smooth operation and device compatibility, as depicted in Figure \ref{fig:roadwatch}.

Data collection involved vehicles fitted with dashcams and external GPS loggers to obtain high-fidelity imagery and position data on urban and rural routes. Videos were processed at 30 FPS, with OCR-driven routines for privacy-preserving blurring (license plates, faces) and precise timestamp extraction. The YOLOv8 model, finely tuned on a region-specific, custom-annotated region specific dataset, detected potholes with bounding boxes, severity ratings, and supporting grading logic. All detection metadata, including geocoordinates, times, severity, and frame encodings, were streamed through Base64 and persisted in a SQLite/PostgreSQL hybrid database for speed and scalability.

The authorization and secure login portals ensured that only verified personnel could input or update critical road and contract information. The integration of Smart Road Construction Mapper enabled segment-based annotation of road health, warranty, and contractor records directly onto the map. A rule-based road health engine dynamically color coded and monitors road segments based on live detection data and warranty metadata.

Automated governance mechanisms further enhanced the platform: critical thresholds, defect liability deadlines, or abnormal segment deterioration triggered real-time notifications via Twilio, SendGrid, or Firebase to both contractors and road authorities.

By relying on consumer-grade dashcams and GPS, edge-processed results, and widely adopted open source technologies (Django, React, Leaflet, QGIS), \rwn minimizes infrastructure and licensing costs underscoring its practicality for broad deployment. In general, the experiment demonstrates how comprehensive, transparent, and accountable road infrastructure monitoring is achievable using accessible, extensible web and geospatial technology, with active support for smart governance features and community engagement at scale.
\section{Discussion and Analysis}

This section outlines the strengths and weakness encountered during the deployment of the \rwn system.

\subsection{Advantages}

The enhanced \rwn platform overcomes key limitations of earlier approaches by leveraging advanced region-specific YOLOv8 fine-tuning and an integrated Smart Road Construction Mapper, providing greater resilience in diverse Indian road and environmental contexts. This version further establishes a unified, end-to-end ecosystem spanning real-time pothole detection, geo-temporal synchronization, warranty-aware road health analytics, and interactive, authority-authenticated web visualization. The revamped interface, complemented by dynamic segment-based color coding and accessible Base64 image previews, streamlines usability for both administrators and the community while providing essential spatial, temporal, and severity insights directly linked to contractor and warranty data for evidence-driven maintenance.

\rwn further empowers smart governance initiatives through automated tracking of contract compliance, budget optimization, repair prioritization, and warranty enforcement, all mapped to individual road segments. Its public dashboard improves transparency and encourages citizen participation, while automated alerts and repair verification features enable seamless accountability and continuous infrastructure improvement. By facilitating granular, real-time insights and enabling operational transparency at scale, it enables truly data-driven infrastructure policy and resource allocation, advancing beyond the limits of sporadic manual oversight.

\subsection{Challenges}

Several technical hurdles had to be overcome to achieve the functionality and reliability demanded of the pothole detection platform.

\begin{itemize}
    \item \textbf{Model Selection and Compatibility:} Various versions of YOLO (v4, v5, v7, v8) were benchmarked for detection accuracy. YOLOv8 achieved the best results, necessitating migration of associated scripts and adaptation to new library dependencies.
    
   \item \textbf{Domain-Specific Dataset Requirements:} Off-the-shelf YOLO models did not generalize to Indian roads due to domain gaps, including inconsistency in the appearance of the pothole (wet/dry, shallow/deep), regional surface noise (tar patches, oil stains), and diverse exposure conditions. As a remedy, a large custom dataset of manually labeled dashcam images was curated using Roboflow and amplified using data augmentation.

    \item \textbf{Temporal Synchronization Issues:}
    \begin{itemize}
        \item YOLO-based text extraction was initially ineffective due to inconsistent model performance and variable overlay placement.
        \item The solution implemented OCR with extraction of the targeted region of interest (ROI).
        \item Challenges arose from variations in timestamp format, incorrect character recognition (e.g., ':' as '.', '/' as '1'), and non-uniform representation of milliseconds.
        \item These were resolved through normalization and the application of multiple regex-based parsing routines.
    \end{itemize}
    
    \item \textbf{GPS-Video Synchronization:} GPS logs (UTC) and dashcam video timestamps (IST/local) required precise calibration, determined as 5 hours, 30 minutes, 44 seconds offset. Geolocation precision was improved by grounding the coordinates to five decimal places, reducing jitter from GPS noise.
    
    \item \textbf{Spatial Redundancy Management:} The high frame rate (30fps) led to multiple detections of the same pothole, introducing redundant data and duplicate markers. Redundancy was minimized using a Haversine-based clustering approach with a 2.5 meter threshold, implemented via the geopy.distance library.
    
    \item \textbf{Image Storage and Retrieval Optimization:} Directly storing image frames in the database led to slowdowns and JSON compatibility issues. Encoding images as Base64 strings allowed efficient storage, API compatibility, and eliminated the need for a dedicated file system.
    
    \item \textbf{Route Matching Accuracy:} The OSRM routing service could generate multiple valid routes between identical endpoints, requiring alignment and periodic recalibration against the actual OSM data.
    
   \textbf{Route Matching Accuracy:} Initially, road segments were selected as straight lines connecting start and end points. To reflect the actual road path, we utilized the Open Source Routing Machine (OSRM) to derive the most probable route. However, caution is still required because different types of roads, such as highways with multiple lanes and standard roads, can visually resemble each other when viewed at a zoomed-out scale. Therefore, we continuously validate and refine route matching to ensure accurate segmentation, especially in complex networks with overlapping or parallel roads.
    
    \item \textbf{Data Redundancy:} Repeated user contributions along the same road could result in segment duplication, necessitating advanced deduplication similar to that used in spatial clustering.
    
    \item \textbf{Verification Reliability:} Confirmation that repairs are completed depends on receiving new drive data after repair, which may be delayed if vehicles infrequently traverse the segment.

\end{itemize}

\section{Conclusion and Future Work}

This work presents \rwn, an advanced end-to-end framework for automated pothole detection, geotagging mapping, and road health governance through the integration of deep learning, based on YOLO, OCR-driven timestamp extraction, and synchronized GPS data, designed for the complexities of Indian road environments and intelligent road governance. The integration of "Interactive Road Contract Mapping and Accountability" bridges the gap between real-time road conditions and official contract data, enabling warranty-aware monitoring, automated notifications, and transparent accountability. The open and interactive interface of the OpenStreetMap based platform empowers both authorities and citizens to visualize, verify, and act on the insights of the infrastructure in real time. In general, \rwn demonstrates how the fusion of computer vision, geospatial intelligence, and web-based governance can transform traditional road maintenance into a data-driven, transparent and participatory process, advancing the goal of sustainable and intelligent transportation infrastructure for smart cities.

Future work will extend \rwn beyond the detection of potholes to a comprehensive assessment of road quality and smart city integration. The system will incorporate algorithms for pavement distress classification, surface roughness estimation, and waterlogging detection, complemented by standardized indices such as the International Roughness Index (IRI) for quantitative evaluation. Advancements in computer vision, including 3D reconstruction for depth estimation and multi-modal fusion with LiDAR or inertial sensors, will enhance detection precision and structural insight. Furthermore, integration with traffic management, IoT sensor networks, and autonomous vehicle systems will position iWatchRoadv2 as a core component of intelligent transportation and urban infrastructure management.

\begin{acks}
This work has benefited from the use of AI language tools (LLMs) for non-substantive editing, including phrasing and grammatical correction. All intellectual content is original and authored by the contributors.
\end{acks}

\bibliographystyle{ACM-Reference-Format}
\bibliography{main}


\begin{thebibliography}{29}


\ifx \showCODEN    \undefined \def \showCODEN     #1{\unskip}     \fi
\ifx \showISBNx    \undefined \def \showISBNx     #1{\unskip}     \fi
\ifx \showISBNxiii \undefined \def \showISBNxiii  #1{\unskip}     \fi
\ifx \showISSN     \undefined \def \showISSN      #1{\unskip}     \fi
\ifx \showLCCN     \undefined \def \showLCCN      #1{\unskip}     \fi
\ifx \shownote     \undefined \def \shownote      #1{#1}          \fi
\ifx \showarticletitle \undefined \def \showarticletitle #1{#1}   \fi
\ifx \showURL      \undefined \def \showURL       {\relax}        \fi
\providecommand\bibfield[2]{#2}
\providecommand\bibinfo[2]{#2}
\providecommand\natexlab[1]{#1}
\providecommand\showeprint[2][]{arXiv:#2}

\bibitem[AI(2020)]%
        {easyocr}
\bibfield{author}{\bibinfo{person}{Jaided AI}.} \bibinfo{year}{2020}\natexlab{}.
\newblock \bibinfo{title}{EasyOCR: Ready-to-use OCR with 80+ languages supported}.
\newblock \bibinfo{howpublished}{\url{https://github.com/JaidedAI/EasyOCR}}.
\newblock


\bibitem[Anthopoulos and Others(2024)]%
        {anthopoulos2024webandthecity}
\bibfield{author}{\bibinfo{person}{Loukas~G. Anthopoulos} {and} \bibinfo{person}{Others}.} \bibinfo{year}{2024}\natexlab{}.
\newblock \showarticletitle{WebAndTheCity: Workshops on Smart Cities and Responsible Web}.
\newblock \bibinfo{journal}{\emph{ACM Web Conference Proceedings}} (\bibinfo{year}{2024}).
\newblock
\urldef\tempurl%
\url{https://dl.acm.org/doi/10.1145/3589335.3641293}
\showURL{%
\tempurl}


\bibitem[Barrón et~al\mbox{.}(2014)]%
        {6975489}
\bibfield{author}{\bibinfo{person}{José Pablo~Gómez Barrón}, \bibinfo{person}{Miguel~Ángel Manso}, \bibinfo{person}{Ramón Alcarria}, {and} \bibinfo{person}{Rufino Pérez~Gómez}.} \bibinfo{year}{2014}\natexlab{}.
\newblock \showarticletitle{A Mobile Crowdsourcing Platform for Urban Infrastructure Maintenance}. In \bibinfo{booktitle}{\emph{2014 Eighth International Conference on Innovative Mobile and Internet Services in Ubiquitous Computing}}. \bibinfo{pages}{358--363}.
\newblock
\href{https://doi.org/10.1109/IMIS.2014.49}{doi:\nolinkurl{10.1109/IMIS.2014.49}}


\bibitem[Carrera et~al\mbox{.}(2013)]%
        {Carrera2013StreetBump}
\bibfield{author}{\bibinfo{person}{F. Carrera}, \bibinfo{person}{S. Guerin}, {and} \bibinfo{person}{J.~B. Thorp}.} \bibinfo{year}{2013}\natexlab{}.
\newblock \showarticletitle{By the People, for the People: The Crowdsourcing of "StreetBump": An Automatic Pothole Mapping App}.
\newblock \bibinfo{journal}{\emph{International Archives of the Photogrammetry, Remote Sensing and Spatial Information Sciences}}  \bibinfo{volume}{XL-4/W1} (\bibinfo{year}{2013}), \bibinfo{pages}{19--23}.
\newblock
\href{https://doi.org/10.5194/isprsarchives-XL-4-W1-19-2013}{doi:\nolinkurl{10.5194/isprsarchives-XL-4-W1-19-2013}}


\bibitem[Chowdary et~al\mbox{.}(2025)]%
        {11031494}
\bibfield{author}{\bibinfo{person}{S.~Kranthi~Kumar Chowdary}, \bibinfo{person}{Y. Harshith}, {and} \bibinfo{person}{T. Preethiya}.} \bibinfo{year}{2025}\natexlab{}.
\newblock \showarticletitle{Smart Pothole Detection and Traffic Sign Identification for Indian Roads: A Machine Learning Approach Using Yolov11}. In \bibinfo{booktitle}{\emph{2025 International Conference on Data Science and Business Systems (ICDSBS)}}. \bibinfo{pages}{1--6}.
\newblock
\href{https://doi.org/10.1109/ICDSBS63635.2025.11031494}{doi:\nolinkurl{10.1109/ICDSBS63635.2025.11031494}}


\bibitem[Doan et~al\mbox{.}(2011)]%
        {doan2011crowdsourcing}
\bibfield{author}{\bibinfo{person}{AnHai Doan}, \bibinfo{person}{Raghu Ramakrishnan}, {and} \bibinfo{person}{Alon~Y. Halevy}.} \bibinfo{year}{2011}\natexlab{}.
\newblock \showarticletitle{Crowdsourcing Systems on the World-Wide Web}.
\newblock \bibinfo{journal}{\emph{Commun. ACM}} \bibinfo{volume}{54}, \bibinfo{number}{4} (\bibinfo{date}{April} \bibinfo{year}{2011}), \bibinfo{pages}{86--96}.
\newblock
\href{https://doi.org/10.1145/1924421.1924442}{doi:\nolinkurl{10.1145/1924421.1924442}}


\bibitem[Dwyer et~al\mbox{.}(2025)]%
        {dwyer2025roboflow}
\bibfield{author}{\bibinfo{person}{Brad Dwyer}, \bibinfo{person}{Joseph Nelson}, \bibinfo{person}{Tom Hansen}, {et~al\mbox{.}}} \bibinfo{year}{2025}\natexlab{}.
\newblock \bibinfo{booktitle}{\emph{Roboflow (Version 1.0)}}.
\newblock
\urldef\tempurl%
\url{https://roboflow.com}
\showURL{%
\tempurl}
\newblock
\shownote{Computer vision platform}.


\bibitem[Fan et~al\mbox{.}(2018)]%
        {8577119}
\bibfield{author}{\bibinfo{person}{Rui Fan}, \bibinfo{person}{Yanan Liu}, \bibinfo{person}{Xingrui Yang}, \bibinfo{person}{Mohammud~Junaid Bocus}, \bibinfo{person}{Naim Dahnoun}, {and} \bibinfo{person}{Scott Tancock}.} \bibinfo{year}{2018}\natexlab{}.
\newblock \showarticletitle{Real-Time Stereo Vision for Road Surface 3-D Reconstruction}. In \bibinfo{booktitle}{\emph{2018 IEEE International Conference on Imaging Systems and Techniques (IST)}}. \bibinfo{pages}{1--6}.
\newblock
\href{https://doi.org/10.1109/IST.2018.8577119}{doi:\nolinkurl{10.1109/IST.2018.8577119}}


\bibitem[Hoseini et~al\mbox{.}(2023)]%
        {hoseini2023pothole}
\bibfield{author}{\bibinfo{person}{M. Hoseini}, \bibinfo{person}{S. Puliti}, \bibinfo{person}{S. Hoffmann}, {and} \bibinfo{person}{R. Astrup}.} \bibinfo{year}{2023}\natexlab{}.
\newblock \showarticletitle{Pothole detection in the woods: a deep learning approach for forest road surface monitoring with dashcams}.
\newblock \bibinfo{journal}{\emph{International Journal of Forest Engineering}} \bibinfo{volume}{35}, \bibinfo{number}{2} (\bibinfo{year}{2023}), \bibinfo{pages}{303--312}.
\newblock
\href{https://doi.org/10.1080/14942119.2023.2290795}{doi:\nolinkurl{10.1080/14942119.2023.2290795}}


\bibitem[Huang et~al\mbox{.}(2016)]%
        {10.1145/2983323.2983886}
\bibfield{author}{\bibinfo{person}{Chao Huang}, \bibinfo{person}{Xian Wu}, {and} \bibinfo{person}{Dong Wang}.} \bibinfo{year}{2016}\natexlab{}.
\newblock \showarticletitle{Crowdsourcing-based Urban Anomaly Prediction System for Smart Cities}. In \bibinfo{booktitle}{\emph{Proceedings of the 25th ACM International on Conference on Information and Knowledge Management}} (Indianapolis, Indiana, USA) \emph{(\bibinfo{series}{CIKM '16})}. \bibinfo{publisher}{Association for Computing Machinery}, \bibinfo{address}{New York, NY, USA}, \bibinfo{pages}{1969–1972}.
\newblock
\showISBNx{9781450340731}
\href{https://doi.org/10.1145/2983323.2983886}{doi:\nolinkurl{10.1145/2983323.2983886}}


\bibitem[J et~al\mbox{.}(2020)]%
        {9112424}
\bibfield{author}{\bibinfo{person}{Dharneeshkar J}, \bibinfo{person}{Soban~Dhakshana V}, \bibinfo{person}{Aniruthan S~A}, \bibinfo{person}{Karthika R}, {and} \bibinfo{person}{Latha Parameswaran}.} \bibinfo{year}{2020}\natexlab{}.
\newblock \showarticletitle{Deep Learning based Detection of potholes in Indian roads using YOLO}. In \bibinfo{booktitle}{\emph{2020 International Conference on Inventive Computation Technologies (ICICT)}}. \bibinfo{pages}{381--385}.
\newblock
\href{https://doi.org/10.1109/ICICT48043.2020.9112424}{doi:\nolinkurl{10.1109/ICICT48043.2020.9112424}}


\bibitem[Jocher et~al\mbox{.}(2023)]%
        {yolov8_ultralytics}
\bibfield{author}{\bibinfo{person}{Glenn Jocher}, \bibinfo{person}{Ayush Chaurasia}, {and} \bibinfo{person}{Jing Qiu}.} \bibinfo{year}{2023}\natexlab{}.
\newblock \bibinfo{booktitle}{\emph{Ultralytics YOLOv8}}.
\newblock
\urldef\tempurl%
\url{https://github.com/ultralytics/ultralytics}
\showURL{%
\tempurl}


\bibitem[Lincy et~al\mbox{.}(2023)]%
        {Lincy}
\bibfield{author}{\bibinfo{person}{A. Lincy}, \bibinfo{person}{G. Dhanarajan}, \bibinfo{person}{S. Sanjay~Kumar}, {and} \bibinfo{person}{B. Gobinath}.} \bibinfo{year}{2023}\natexlab{}.
\newblock \showarticletitle{Road Pothole Detection System}.
\newblock \bibinfo{journal}{\emph{ITM Web Conf.}}  \bibinfo{volume}{53} (\bibinfo{year}{2023}), \bibinfo{pages}{01008}.
\newblock
\href{https://doi.org/10.1051/itmconf/20235301008}{doi:\nolinkurl{10.1051/itmconf/20235301008}}


\bibitem[Ma et~al\mbox{.}(2022)]%
        {Ma_2022}
\bibfield{author}{\bibinfo{person}{Nachuan Ma}, \bibinfo{person}{Jiahe Fan}, \bibinfo{person}{Wenshuo Wang}, \bibinfo{person}{Jin Wu}, \bibinfo{person}{Yu Jiang}, \bibinfo{person}{Lihua Xie}, {and} \bibinfo{person}{Rui Fan}.} \bibinfo{year}{2022}\natexlab{}.
\newblock \showarticletitle{Computer vision for road imaging and pothole detection: a state-of-the-art review of systems and algorithms}.
\newblock \bibinfo{journal}{\emph{Transportation Safety and Environment}} \bibinfo{volume}{4}, \bibinfo{number}{4} (\bibinfo{date}{Nov.} \bibinfo{year}{2022}).
\newblock
\showISSN{2631-4428}
\href{https://doi.org/10.1093/tse/tdac026}{doi:\nolinkurl{10.1093/tse/tdac026}}


\bibitem[mysociety et~al\mbox{.}(2024)]%
        {fixmystreet2024}
\bibfield{author}{\bibinfo{person}{mysociety} {et~al\mbox{.}}} \bibinfo{year}{2024}\natexlab{}.
\newblock \showarticletitle{FixMyStreet: Crowdsourcing street problem reports}.
\newblock  (\bibinfo{year}{2024}).
\newblock
\urldef\tempurl%
\url{https://fixmystreet.org}
\showURL{%
\tempurl}


\bibitem[Omar and Kumar(2024)]%
        {10.1007/s42979-024-02887-1}
\bibfield{author}{\bibinfo{person}{Mohd Omar} {and} \bibinfo{person}{Pradeep Kumar}.} \bibinfo{year}{2024}\natexlab{}.
\newblock \showarticletitle{PD-ITS: Pothole Detection Using YOLO Variants for Intelligent Transport System}.
\newblock \bibinfo{journal}{\emph{SN Comput. Sci.}} \bibinfo{volume}{5}, \bibinfo{number}{5} (\bibinfo{date}{May} \bibinfo{year}{2024}), \bibinfo{numpages}{16}~pages.
\newblock
\href{https://doi.org/10.1007/s42979-024-02887-1}{doi:\nolinkurl{10.1007/s42979-024-02887-1}}


\bibitem[{OpenStreetMap contributors}(2024)]%
        {openstreetmap}
\bibfield{author}{\bibinfo{person}{{OpenStreetMap contributors}}.} \bibinfo{year}{2024}\natexlab{}.
\newblock \bibinfo{title}{Planet dump retrieved from https://planet.openstreetmap.org}.
\newblock \bibinfo{howpublished}{\url{https://www.openstreetmap.org}}.
\newblock


\bibitem[Qiu et~al\mbox{.}(2019)]%
        {qiu2019crowdmapping}
\bibfield{author}{\bibinfo{person}{Shengyuan Qiu}, \bibinfo{person}{Achilleas Psyllidis}, \bibinfo{person}{Alessandro Bozzon}, {and} \bibinfo{person}{Geert-Jan Houben}.} \bibinfo{year}{2019}\natexlab{}.
\newblock \showarticletitle{Crowd-Mapping Urban Objects from Street-Level Imagery}. In \bibinfo{booktitle}{\emph{The World Wide Web Conference}} (San Francisco, CA, USA) \emph{(\bibinfo{series}{WWW '19})}. \bibinfo{publisher}{ACM}, \bibinfo{address}{New York, NY, USA}, \bibinfo{pages}{1521--1531}.
\newblock
\showISBNx{9781450366748}
\href{https://doi.org/10.1145/3308558.3313651}{doi:\nolinkurl{10.1145/3308558.3313651}}


\bibitem[Rasyid et~al\mbox{.}(2019)]%
        {8901626}
\bibfield{author}{\bibinfo{person}{Alfandino Rasyid}, \bibinfo{person}{Mochammad~Rifki Ulil~Albaab}, \bibinfo{person}{Muhammad Fajrul~Falah}, \bibinfo{person}{Yohanes~Yohanie Fridelin~Panduman}, \bibinfo{person}{Alviansyah Arman~Yusuf}, \bibinfo{person}{Dwi~Kurnia Basuki}, \bibinfo{person}{Anang Tjahjono}, \bibinfo{person}{Rizqi~Putri Nourma~Budiarti}, \bibinfo{person}{Sritrusta Sukaridhoto}, \bibinfo{person}{Firman Yudianto}, {and} \bibinfo{person}{Hendro Wicaksono}.} \bibinfo{year}{2019}\natexlab{}.
\newblock \showarticletitle{Pothole Visual Detection using Machine Learning Method integrated with Internet of Thing Video Streaming Platform}. In \bibinfo{booktitle}{\emph{2019 International Electronics Symposium (IES)}}. \bibinfo{pages}{672--675}.
\newblock
\href{https://doi.org/10.1109/ELECSYM.2019.8901626}{doi:\nolinkurl{10.1109/ELECSYM.2019.8901626}}


\bibitem[Redmon et~al\mbox{.}(2016)]%
        {redmon2016you}
\bibfield{author}{\bibinfo{person}{Joseph Redmon}, \bibinfo{person}{Santosh Divvala}, \bibinfo{person}{Ross Girshick}, {and} \bibinfo{person}{Ali Farhadi}.} \bibinfo{year}{2016}\natexlab{}.
\newblock \showarticletitle{You only look once: Unified, real-time object detection}. In \bibinfo{booktitle}{\emph{Proceedings of the IEEE conference on computer vision and pattern recognition}}. \bibinfo{pages}{779--788}.
\newblock


\bibitem[Safyari et~al\mbox{.}(2024)]%
        {Safyari2024}
\bibfield{author}{\bibinfo{person}{Yeganeh Safyari}, \bibinfo{person}{Masoud Mahdianpari}, {and} \bibinfo{person}{Hamid Shiri}.} \bibinfo{year}{2024}\natexlab{}.
\newblock \showarticletitle{A Review of Vision-Based Pothole Detection Methods Using Computer Vision and Machine Learning}.
\newblock \bibinfo{journal}{\emph{Sensors}} \bibinfo{volume}{24}, \bibinfo{number}{17} (\bibinfo{year}{2024}), \bibinfo{pages}{5652}.
\newblock
\href{https://doi.org/10.3390/s24175652}{doi:\nolinkurl{10.3390/s24175652}}


\bibitem[Sahoo et~al\mbox{.}(2025)]%
        {sahoo2025iwatchroadscalabledetectiongeospatial}
\bibfield{author}{\bibinfo{person}{Rishi~Raj Sahoo}, \bibinfo{person}{Surbhi~Saswati Mohanty}, {and} \bibinfo{person}{Subhankar Mishra}.} \bibinfo{year}{2025}\natexlab{}.
\newblock \bibinfo{title}{iWatchRoad: Scalable Detection and Geospatial Visualization of Potholes for Smart Cities}.
\newblock
\showeprint[arxiv]{2508.10945}~[cs.CV]
\urldef\tempurl%
\url{https://arxiv.org/abs/2508.10945}
\showURL{%
\tempurl}


\bibitem[Sminage et~al\mbox{.}(2025)]%
        {11042541}
\bibfield{author}{\bibinfo{person}{Amxson Sminage}, \bibinfo{person}{Delvin~P B}, \bibinfo{person}{Derick Davies}, \bibinfo{person}{Vivek~K J}, {and} \bibinfo{person}{Jasmy Davies}.} \bibinfo{year}{2025}\natexlab{}.
\newblock \showarticletitle{SafeDrive: Intelligent Pothole Detection and Mapping System}. In \bibinfo{booktitle}{\emph{2025 2nd International Conference on Trends in Engineering Systems and Technologies (ICTEST)}}, Vol.~\bibinfo{volume}{1}. \bibinfo{pages}{1--6}.
\newblock
\href{https://doi.org/10.1109/ICTEST64710.2025.11042541}{doi:\nolinkurl{10.1109/ICTEST64710.2025.11042541}}


\bibitem[team(2024)]%
        {seeclickfix2024}
\bibfield{author}{\bibinfo{person}{SeeClickFix team}.} \bibinfo{year}{2024}\natexlab{}.
\newblock \showarticletitle{SeeClickFix: A platform for civic issue reporting}.
\newblock  (\bibinfo{year}{2024}).
\newblock
\urldef\tempurl%
\url{https://seeclickfix.com}
\showURL{%
\tempurl}


\bibitem[Tempelmeier and Demidova(2022)]%
        {Tempelmeier_2022}
\bibfield{author}{\bibinfo{person}{Nicolas Tempelmeier} {and} \bibinfo{person}{Elena Demidova}.} \bibinfo{year}{2022}\natexlab{}.
\newblock \showarticletitle{Attention-Based Vandalism Detection in OpenStreetMap}. In \bibinfo{booktitle}{\emph{Proceedings of the ACM Web Conference 2022}} \emph{(\bibinfo{series}{WWW ’22})}. \bibinfo{publisher}{ACM}, \bibinfo{pages}{643–651}.
\newblock
\href{https://doi.org/10.1145/3485447.3512224}{doi:\nolinkurl{10.1145/3485447.3512224}}


\bibitem[Truong et~al\mbox{.}(2018)]%
        {truong_et_al:LIPIcs.GISCIENCE.2018.61}
\bibfield{author}{\bibinfo{person}{Quy~Thy Truong}, \bibinfo{person}{Guillaume Touya}, {and} \bibinfo{person}{Cyril de Runz}.} \bibinfo{year}{2018}\natexlab{}.
\newblock \showarticletitle{{Towards Vandalism Detection in OpenStreetMap Through a Data Driven Approach}}. In \bibinfo{booktitle}{\emph{10th International Conference on Geographic Information Science (GIScience 2018)}} \emph{(\bibinfo{series}{Leibniz International Proceedings in Informatics (LIPIcs)}, Vol.~\bibinfo{volume}{114})}, \bibfield{editor}{\bibinfo{person}{Stephan Winter}, \bibinfo{person}{Amy Griffin}, {and} \bibinfo{person}{Monika Sester}} (Eds.). \bibinfo{publisher}{Schloss Dagstuhl -- Leibniz-Zentrum f{\"u}r Informatik}, \bibinfo{address}{Dagstuhl, Germany}, \bibinfo{pages}{61:1--61:7}.
\newblock
\showISBNx{978-3-95977-083-5}
\showISSN{1868-8969}
\href{https://doi.org/10.4230/LIPIcs.GISCIENCE.2018.61}{doi:\nolinkurl{10.4230/LIPIcs.GISCIENCE.2018.61}}


\bibitem[Vahdat-Nejad et~al\mbox{.}(2022)]%
        {vahdatnejad2022surveycrowdsourcingapplicationssmart}
\bibfield{author}{\bibinfo{person}{Hamed Vahdat-Nejad}, \bibinfo{person}{Tahereh Tamadon}, \bibinfo{person}{Fatemeh Salmani}, \bibinfo{person}{Zeynab kiani Zadegan}, \bibinfo{person}{Sajedeh Abbasi}, {and} \bibinfo{person}{Fateme-Sadat Seyyedi}.} \bibinfo{year}{2022}\natexlab{}.
\newblock \bibinfo{title}{A Survey on Crowdsourcing Applications in Smart Cities}.
\newblock
\showeprint[arxiv]{2204.05421}~[cs.HC]
\urldef\tempurl%
\url{https://arxiv.org/abs/2204.05421}
\showURL{%
\tempurl}


\bibitem[Wang et~al\mbox{.}(2016)]%
        {wang2016publicsense}
\bibfield{author}{\bibinfo{person}{Zeyu Wang}, \bibinfo{person}{Bin Guo}, \bibinfo{person}{Zhiwen Yu}, \bibinfo{person}{Wei Wu}, \bibinfo{person}{Jingmin Zhang}, \bibinfo{person}{Zhu Wang}, {and} \bibinfo{person}{Huihui Chen}.} \bibinfo{year}{2016}\natexlab{}.
\newblock \showarticletitle{PublicSense: A Crowd Sensing Platform for Public Facility Management in Smart Cities}. In \bibinfo{booktitle}{\emph{2016 Intl IEEE Conferences on Ubiquitous Intelligence \& Computing, Advanced and Trusted Computing, Scalable Computing and Communications, Cloud and Big Data Computing, Internet of People, and Smart World Congress (UIC/ATC/ScalCom/CBDCom/IoP/SmartWorld)}}. \bibinfo{publisher}{IEEE}, \bibinfo{address}{Toulouse, France}, \bibinfo{pages}{440--447}.
\newblock
\href{https://doi.org/10.1109/UIC-ATC-ScalCom-CBDCom-IoP-SmartWorld.2016.0078}{doi:\nolinkurl{10.1109/UIC-ATC-ScalCom-CBDCom-IoP-SmartWorld.2016.0078}}


\bibitem[Yebes et~al\mbox{.}(2021)]%
        {Yebes_2021}
\bibfield{author}{\bibinfo{person}{J.~Javier Yebes}, \bibinfo{person}{David Montero}, {and} \bibinfo{person}{Ignacio Arriola}.} \bibinfo{year}{2021}\natexlab{}.
\newblock \showarticletitle{Learning to Automatically Catch Potholes in Worldwide Road Scene Images}.
\newblock \bibinfo{journal}{\emph{IEEE Intelligent Transportation Systems Magazine}} \bibinfo{volume}{13}, \bibinfo{number}{3} (\bibinfo{year}{2021}), \bibinfo{pages}{192–205}.
\newblock
\showISSN{1941-1197}
\href{https://doi.org/10.1109/mits.2019.2926370}{doi:\nolinkurl{10.1109/mits.2019.2926370}}


\end{thebibliography}



\end{document}